\documentclass[journal]{IEEEtran}
\usepackage[utf8]{inputenc}
\usepackage[pdftex]{graphicx}
\usepackage{tikzducks}
\usepackage{amsmath}
\usepackage{amssymb}
\usepackage{xcolor}
\usepackage{extarrows}
\usepackage{algorithmic}
\usepackage{stfloats}
\usepackage{url}
\usepackage{cases}
\usepackage{hyperref} % 引入 hyperref 包以支持超链接
\usepackage{textgreek} % Support for Greek letters in text
\usepackage{verbatim} % for comment environment
\usepackage{lipsum} % placeholder
\hypersetup{
    colorlinks=true, % 启用链接颜色
    linkcolor=blue,  % 内部链接颜色
    urlcolor=magenta    % 网址颜色
}
\definecolor{darkblue}{RGB}{0, 84, 180} % RGB 值为

\begin{document}
%
% \title{SuckTac: Camera-based Tactile Sucker \\ for Intelligent Surface Perception, Hight Terrain Adaptiveness and Manipulations}
\title{SuckTac: Camera-based Tactile Sucker \\ for Unstructured Surface Perception and Interaction}
%\author{
%Anonymous Authors
\author{Ruiyong Yuan$^{\ast}$, Jieji Ren$^{\ast}$, Zhanxuan Peng, Feifei Chen$^{**}$, Guoying Gu$^{**}$\\

 \thanks{$^\ast$Equally Contributed; $^{**}$Corresponding Author.}
 \thanks{Ruiyong Yuan, Jieji Ren, Zhanxuan Peng, Feifei Chen and Guoying Gu are with the School of Mechanical Enginnering, Shanghai Jiao Tong University, Shanghai, China, 200240, China. E-mails: jiejiren, guguoying@sjtu.edu.cn.}
 \thanks{Manuscript received Jun xx, 2025; revised Aug xx, 2025.}
}

% \markboth{Journal of \LaTeX\ Class Files,~Vol.~xx, No.~xx, August~202x}%
% {Shell \MakeLowercase{\textit{et al.}}: Bare Demo of IEEEtran.cls for IEEE Journals}

\maketitle

\begin{abstract}
%\TR{Finished, please check! then remove this!}
%Suckers are significant for robots on fast picking, transferring, manipulation and locomotion on diverse surfaces.
%However, most of the existing suckers lack the high-fidelity perceptual and tactile sensing, which impeding them to resolve the fine-grained geometric features and interaction status of target surface. This limits their robust suction capability in irregular objects and complex, unstructured environments.
%Inspired by the adaptiveness structure and high-performance sensory capabilities of cephalopod suckers, in this paper, we proposed a novel, intelligent sucker, named SuckTac, that integrates the camera-based tactile sensor directly within its optimized structure to providing high-density perception and robust suction. 
%Specifically, through a joint structure design and optimization and based on multi-material integrated casting technique, a camera and light source are embedded into sucker, which enable in-situ, high-density perception of fine details like surface shape, texture and roughness.
%To further enhance robustness and adaptability, the sucker's mechanical design was also optimized by refining its profile, adding a compliant skirt, and incorporating surface microstructure. 
%Extensive experiments, including challenging tasks such as robotic cloth manipulation and soft mobile robot inspection, demonstrate the superior performance and broad applicability of the proposed system. 
% Dataset and experiment details can be found at \href{http://github.com/tacxles/FringTac}{http://github.com/tacxles/FringTac}\GQY{needs revision}.
Suckers are significant for robots in picking, transferring, manipulation and locomotion on diverse surfaces.
However, most of the existing suckers lack high-fidelity perceptual and tactile sensing, which impedes them from resolving the fine-grained geometric features and interaction status of the target surface. This limits their robust performance with irregular objects and in complex, unstructured environments.
Inspired by the adaptive structure and high-performance sensory capabilities of cephalopod suckers, in this paper, we propose a novel, intelligent sucker, named SuckTac, that integrates a camera-based tactile sensor directly within its optimized structure to provide high-density perception and robust suction.
Specifically, through joint structure design and optimization and based on a multi-material integrated casting technique, a camera and light source are embedded into the sucker, which enables in-situ, high-density perception of fine details like surface shape, texture and roughness.
To further enhance robustness and adaptability, the sucker's mechanical design is also optimized by refining its profile, adding a compliant lip, and incorporating surface microstructure.
Extensive experiments, including challenging tasks such as robotic cloth manipulation and soft mobile robot inspection, demonstrate the superior performance and broad applicability of the proposed system.
\end{abstract}


% \fi
\IEEEpeerreviewmaketitle

\section{Introduction}
\begin{figure}[!ht]
    \centering
    \includegraphics[width=\linewidth]{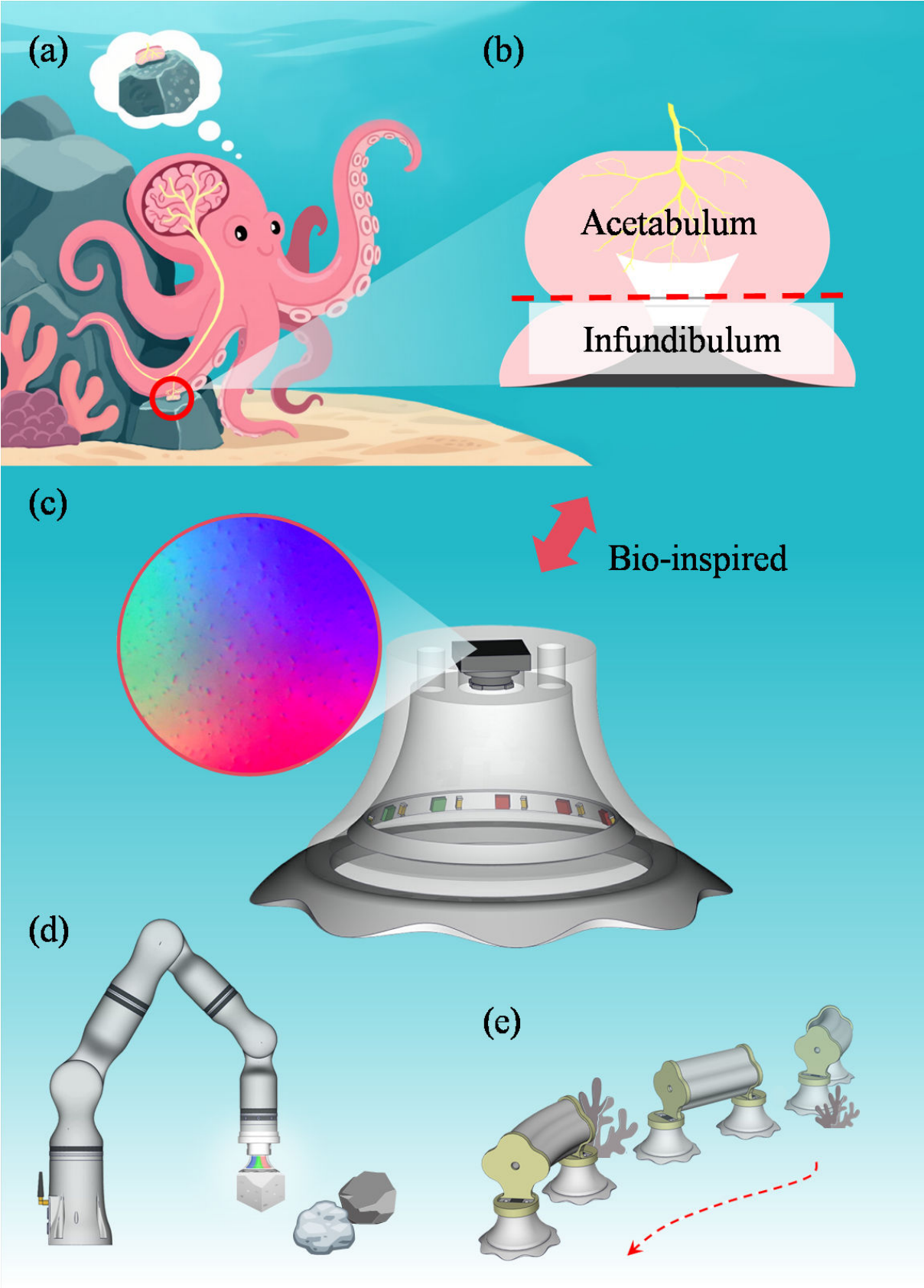}
    \caption{Bionic camera-based tactile sucker and its biological inspiration and robotic applications: (a) Octopus utilize suckers to feel the surface condition of the target; (b) Enlarged view of the natural sucker structure; (c) The proposed bionic camera-based tactile sucker can support robot arm to classify the object and picking as (d), and assist the path planing for soft robot crawling in (e).
    % (d) Robotic arm performing a grasping task with the sucker; (e) Robot equipped with the sucker for crawling.
    }
    \label{fig:teaser}
\end{figure}

%\IEEEPARstart{S}{ucker} sucker is one of the most critical end effector for robots with highly redundant, nondestructive, and compliant interaction~\cite{},
%and facilitate diverse applications, such as objects picking~\cite{}, tools manipulation~\cite{} and claimb locomotion assistance~\cite{}.
%However, their performance depends on a tight seal and a smooth surface, making it difficult to stick to uneven or discontinuous objects. For a robot to plan its grasping strategy and achieve a robust hold, its' sucker need to perceive the state of the target surface and make an effective decision. It is important to equip the sucker with high-performance sensing and touch ability.

\IEEEPARstart{S}{ucker} is one of the most critical end effectors for robots with highly redundant, nondestructive, and compliant interaction~\cite{cupreview},
and facilitates diverse applications, such as object picking~\cite{RN98,RN99,RN102,RN109}, tool manipulation~\cite{RN101,RN112} and climbing locomotion assistance~\cite{RN2,RN100}.
However, its performance depends on a tight seal and a smooth surface, making it difficult to stick to uneven or discontinuous objects. For a robot to plan its grasping strategy and achieve a robust hold, its sucker needs to perceive the state of the target surface and make an effective decision~\cite{RN98,RN99,RN103,RN111,RN107}. It is important to equip the sucker with high-performance sensing and tactile capabilities.

Researchers have explored diverse perception approaches for suckers. Some previous work has introduced pressure perception to estimate suction force and surface roughness, using tools like flow sensors, strain gauges, or LiDAR~\cite{RN99,RN103,RN111}. 
Unfortunately, these methods cannot precisely perceive the target's geometric morphology, which is significant for a seamless seal and stable suction. 
On the other hand, some researchers have introduced vision-based surface detection methods~\cite{zhang2023robot,zhang2022suction} to find the best contact regions~\cite{yoon2024seg2grasp}, but direct observation is strongly interfered by the object appearance, environmental illuminations and optical conditions.
It is difficult for these methods to obtain high-precision and high-density sensing information, such as texture, shape, and other details of the surface, which affects the sucker's mechanical and functional performance, and in turn makes robust locomotion and grasping challenging.

%Inspiring by the suction cup of Cephalopods, which not only has stable, robust adaptive adhesion abilities, but also equipped with high-density and multi-modality sensing capabilities.
%We recognize that high-performance sensing, such as high-density tactile sensing ability, will effectively improve the working ability of the sucker. 
%It is crucial to creatively integrate the high-density tactile sensor into the sucker design.
We draw inspiration from the suction cups of cephalopods, as shown in Fig. \ref{fig:teaser}(a) and (b), which not only possess stable, robust, and adaptive adhesion abilities but are also equipped with high-density, multi-modal sensing capabilities. 
We recognize that high-performance sensing, such as high-density tactile capabilities, can effectively improve the performance of the sucker. Therefore, it is crucial to integrate a high-density tactile sensor into the sucker's design.

Among diverse tactile technology, camera-based tactile sensors utilize the image sensor to observe the soft skin deformation when the robot fingertip interacts with the target surface, and estimate the geometry, contact force and surface friction, which provides a high-density, robust, stable and low-cost tactile approach for robots. 
GelSight is the first to integrate with a robot gripper and provides high-density tactile images for robots~\cite{10122097}, then GelSlim adds markers on the gel surface to estimate contact forces, and enables more compact and integrable designs~\cite{gelslim3.0}.
With fast development, camera-based tactile sensors have become the most popular sensors in the robot community~\cite{zhang2022hardware,fang2025force}, providing robots with abundant tactile information for environmental perception, dexterous grasping, and complex manipulation~\cite{he2025survey}.

This high-performance sensing mechanism provides a potential approach for developing an intelligent sucker with integrated structure-perception capabilities. Recent studies have thus focused on combining camera-based tactile sensing with robotic hands and compliant grippers, such as the GelSight Svelte~\cite{10341646} and the Soft-bubble gripper~\cite{9341534}, as well as integrating the sensor into a pneumatic gripper for a low-cost swab sampling tool~\cite{10342266}.

However, most prior work has focused on integrating these sensors with rigid grippers or fingertips, with limited exploration into their use with soft actuators—particularly the industrially ubiquitous suction cup. A notable exception is an integrated camera-based sensor for a sucker~\cite{RN107}, whose marker-based design could only detect orientation and failed to resolve surface texture and roughness. These factors are critical for predicting and ensuring reliable suction, leaving a significant gap in both research and application.

To address the limitations of classical suckers and advance their use in embodied intelligence, we propose a camera-based tactile sucker named SuckTac, as shown in Fig.~\ref{fig:teaser}(c). This design integrates high-density tactile sensing directly into the sucker's structure and features a jointly optimized configuration for robust suction. Specifically, we use a multi-material integrated casting technique to embed a camera and its light source within the sucker. This provides high-density, in-situ, and detailed tactile perception. To further enhance suction performance, we also improve the sucker's profile, add a lip, and incorporate surface protrusions. By combining integrated perception with actuation, the proposed SuckTac achieves robust, adaptive grasping and locomotion. 
This is demonstrated in its ability to perceive surface texture and roughness, and to handle tasks like robotic clothes grasping and soft mobile robot inspection.
%This is demonstrated in its ability to surface texture and roughness perception, and to handle tasks like robotic clothes grasping and soft mobile robot inspection.

%We believe the proposed SuckTac can improve the performance and application of sucker, as well as expand the potentials and frontier of sucker for research community.
We believe the proposed SuckTac can improve the performance and applications of suckers, as well as expand the potential and frontiers of suckers for the research community.The contributions of this work can be summarized as follows:

\begin{itemize}
    \item Propose a smart camera-based tactile sucker for integrated high-density sensing capability;
    \item Jointly optimize the sucker design structure for improving the suction performance and adaptiveness;
    \item Extensive robot grasping and locomotion tests demonstrate the tactile potential of the proposed SuckTac.
\end{itemize}

The rest of this article is organized as follows. 
%Section \ref{sec:Related Work} reviews related works, focusing on camera-based tactile sensors and reconstruction methods. 
Section \ref{sec:Method} details the design and implementation of the SuckTac system. Section \ref{sec:Experiments} describes experimental evaluations on the performance of SuckTac and presents the potential robotic applications. Finally, Section \ref{sec:Conclusion} concludes this article.

\section{Camera-based Tactile Sucker Design}
\label{sec:Method}
This section introduces the design and optimization process for the proposed SuckTac. First, to integrate high-resolution tactile sensing, we engineer a multi-material composite structure to house the internal camera and illumination components. This design effectively embeds a camera-based tactile sensor within the sucker. Second, we systematically optimize the sucker's geometry to enhance its functional performance in three key areas: the profile is shaped to maximize detachment energy, the lip design is iterated to improve adaptability to uneven surfaces, and a microstructure is added to ensure robust attachment on rough textures. As a result of these enhancements, the SuckTac combines high-performance sensing with powerful suction, demonstrating remarkable adaptability.

\subsection{SuckTac Design and Fabrication}
\label{sec:design}
% Ruiyong: Please write this part first
% Design: 讲清楚我们这个吸盘视触觉的基本结构和机理方面，视触觉方面可以在这里就划清楚
% 可以放我们最早的那个结构图，爆炸开来展示内部结构
% 另一边把机理画一下
\begin{figure}[!t]
    \centering
    \includegraphics[width=\columnwidth]{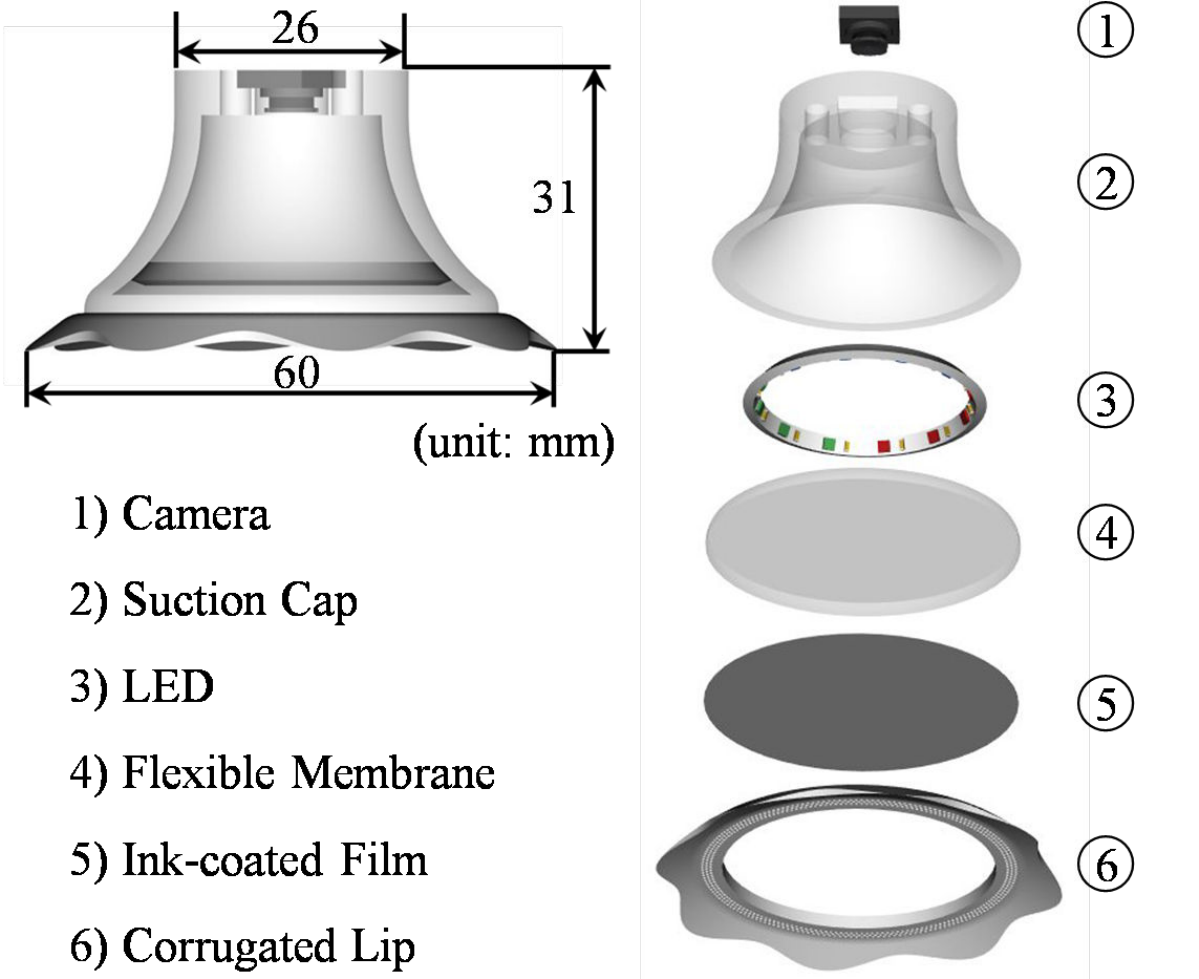}
    \caption{The detailed Structure of SuckTac. The camera-based tactile sensor is artfully integrated into the sucker (Right: Exploded view).}
    \label{fig:structure}
\end{figure}

To enable robust attachment on unstructured surfaces, we introduce a Camera-based Tactile Sucker (SuckTac) that captures high-resolution contact information. The overall structure consists of a suction cap, a flexible membrane, a corrugated lip, and an embedded camera-based tactile sensing module [Fig. \ref{fig:structure}]. The suction cap, inspired by the acetabulum of biological octopus suckers, is fabricated from SYLGARD 184 with high stiffness to provide structural support and prevent collapse or excessive deformation. The flexible membrane, made of transparent Solaris silicone, forms a sealed cavity with the cap, thereby generating negative pressure for suction. The corrugated lip, made of ultra-soft Ecoflex 00-10, mimics the infundibulum to ensure conformal contact and airtight sealing on rough or irregular surfaces.

The camera-based tactile sensing module comprises a compact camera, an RGB LED array for illumination, and an ink-coated film attached to the flexible membrane. When the sucker approaches a target surface, positive pressure is applied inside the cavity, causing the membrane to bulge outward and conform to the surface. This deformation presses the ink-coated film into intimate contact, allowing it to adapt to local irregularities. Micro-scale surface features then induce corresponding deformations in the ink pattern, which are illuminated by the LED array and captured by the camera as high-resolution images. This mechanism allows the system to encode fine-scale surface characteristics, including roughness and micro-texture, into optical tactile information that can be subsequently analyzed for perception and classification. 

To realize the proposed design, we develop a fabrication process consisting of several sequential steps. First, the suction cap is cast in a mold and allowed to cure. After curing, the RGB LED array is bonded to the interior of the cap. Next, The flexible membrane is cast onto the underside of the suction cap, forming the internal cavity. Once the membrane is solidified, the ink-coated film is uniformly applied to the membrane surface via spray coating and allowed to dry. Subsequently, a separately molded corrugated lip is bonded to the base of the membrane using a silicone adhesive. Finally, the compact camera is focused and encapsulated to complete the assembly.

\begin{figure*}[!ht]
    \centering
    \includegraphics[width=\linewidth]{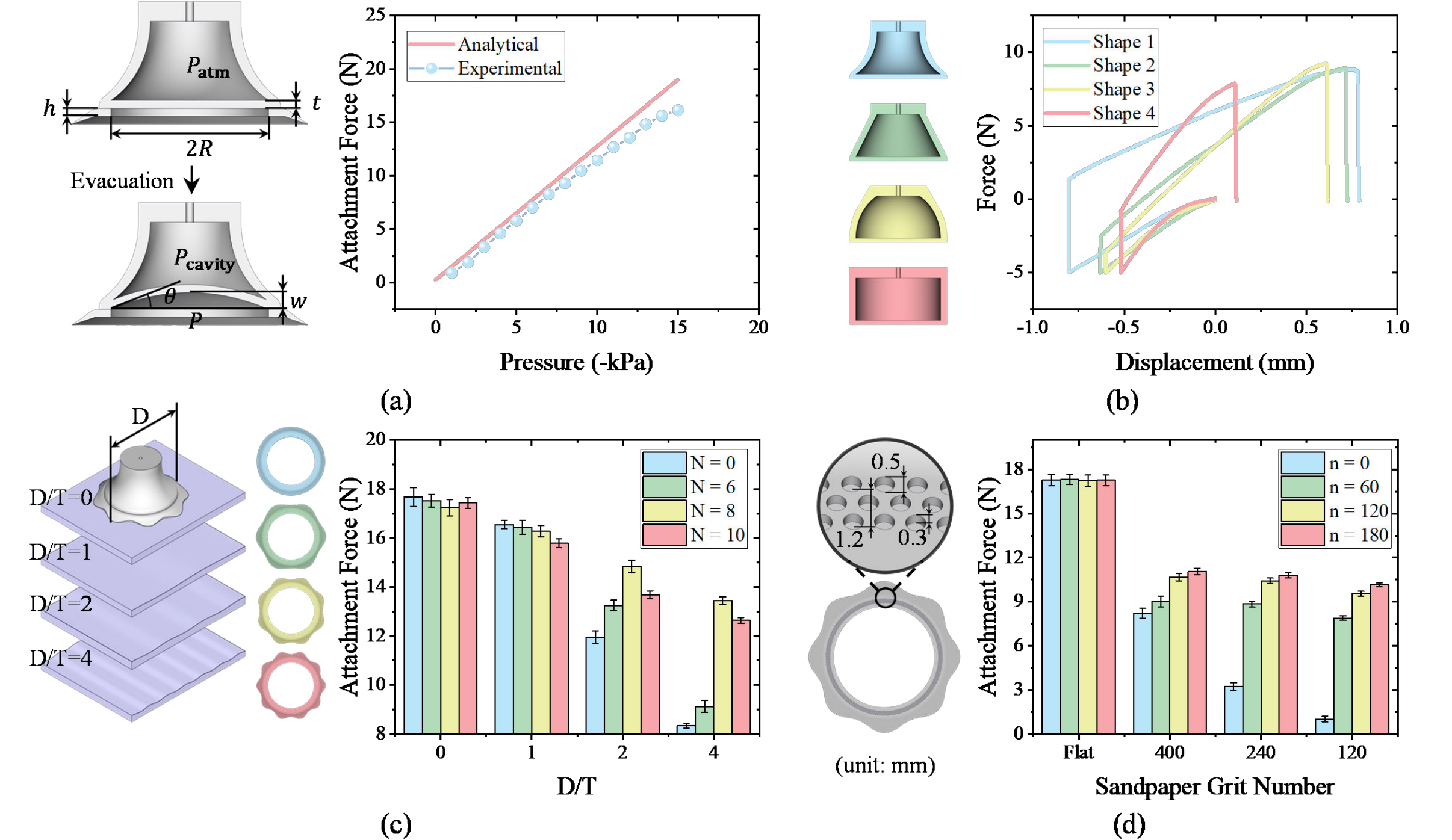}
    \caption{Characterization and performance evaluation of the SuckTac: (a) Modeling of the sucker. (Left) Modeling parameters; (Right) comparison between theoretical predictions and experimental results.(b) Suction force–displacement relationships of four different sucker structures. (c) Suction force comparison of suckers with four lip designs on four types of periodic curved surfaces. (d) Suction force of suckers with different sub-millimeter structure densities tested on sandpapers of varying grit number.}
    \label{fig:improvement}
\end{figure*}

% 这里主要讲视触觉细节的设计，包括相机如何放置、灯带如何设计
% 表面涂层浇筑和设计的工艺
% 喷涂工艺等等细节，
% 总的来说，就是描述如何把吸盘视触觉传感器做出来，也就是我们原理验证时候的技术工艺写一遍；

\subsection{Sucker Modeling}
\label{sec:model}
The sucker modeling mainly focuses on the flexible membrane, since the deformation of the sucker cap can be neglected when the negative pressure is less than $-15$ kPa, which is already sufficient for the intended applications. The membrane is simplified as a circular thin film with radius $R$ and thickness $t$. When a negative pressure $P_{\text{cavity}}$ is applied, the membrane deforms under the pressure difference:
\begin{equation}
\Delta P = P - P_{\text{cavity}},
\label{eq:0}
\end{equation}
where $P$ is the pressure inside the gap between the deformed membrane and the substrate.
The deflection profile of the deformed membrane is approximated by a parabola:
\begin{equation}
z(r) = w \left( 1 - \frac{r^2}{R^2} \right),
\label{eq:1}
\end{equation}
where $w$ is the central deflection.  

The arc length of the deformed membrane $L$ is expressed as:
\begin{equation}
L = 2\int_0^R \sqrt{1 + \left(\frac{\partial z}{\partial r}\right)^2} \, dr.
\label{eq:2}
\end{equation}

The pressure difference $\Delta P$ is balanced by the vertical component of the radial stress $\sigma_r$ at the membrane edge:
\begin{equation}
\pi{R^2}\Delta P = -2\pi{Rt}\sigma_r \sin \theta,
\label{eq:3}
\end{equation}
where $\theta$ is the angle between the tangent at the membrane edge and the horizontal direction. For small deformations,  the sine equals the 
tangent and can be calculated as the derivative of the deflection curve $z(r)$ thus
\begin{equation}
\pi{R^2}\Delta P = -2\pi{Rt}\sigma_r\frac{2w}{R},
\label{eq:4}
\end{equation}

For the thin membrane considered here ($t \ll R$), the axial strain can be assumed negligible, while the circumferential and radial strains are taken to be equal. The Solaris silicone used for the membrane is incompressible, and its hyperelastic behavior is described by the neo-Hookean model. Under these assumptions, the radial stress in the membrane can be calculated as
\begin{equation}
\sigma_r = 2C \left( \left(\frac{L}{2R}\right)^2 - {\left(\frac{2R}{L}\right)^4} \right),
\end{equation}
where $C = 0.098~\text{MPa}$ is the material constant obtained from uniaxial tensile tests. This formulation provides a direct link between the membrane geometry, material properties, and the resulting radial stress.

The internal pressure in the gap between the deformed membrane and the substrate can be estimated using the ideal gas law, which gives
\begin{equation}
P \, \pi R^2 \left(h + \frac{w}{2}\right) = P_{\text{atm}} \, \pi R^2 h,
\end{equation}
where $h$ is the initial height between the undeformed membrane and the substrate.

The resulting attachment force acting on the substrate is then
\begin{equation}
F = \pi R^2 \left(P_{\text{atm}} - P \right).
\label{eq:final}
\end{equation}

By combining (\ref{eq:0})–(\ref{eq:final}), the attachment force $F$ can be expressed explicitly as a function of the applied negative pressure $P_{\text{cavity}}$, providing an analytical link between the actuation input and the resulting adhesion. To assess the validity of this model, we compare its predictions with experimental measurements [Fig.~\ref{fig:improvement}(a)]. The comparison shows good agreement, particularly in the small-deflection regime, demonstrating the accuracy of the proposed model. The parameters used in the analytical calculations are listed in Table~I.

\begin{table}[!t]
\centering
\caption{Parameters used in the analytical model}
\begin{tabular}{c c c c c}
\hline
$R$ (mm) & $t$ (mm) & $C$ (MPa) & $P_{\text{atm}}$ (MPa) & $h$ (mm) \\
\hline
20 & 2 & 0.098 & 0.101 & 2 \\
\hline
\end{tabular}
\end{table}

\subsection{Cross-Sectional Shape Design}
\label{sec:improvement}

In unstructured environments, suckers are prone to disturbances that can compromise attachment stability. However, increasing the flexibility of the sucker has been shown to effectively enhance adhesion reliability. Higher flexibility allows the sucker to better conform to the contact surface during attachment, and also requires greater energy to cause detachment afterward, thereby improving overall contact quality.

To investigate the effect of cross-sectional geometry on attachment, we design four suckers with identical base area and height but different cross-sectional shapes [Fig.~\ref{fig:improvement}(b)]. Experiments are conducted on a smooth acrylic substrate, with each sucker subjected to the same preloading force (5~N) and negative pressure (-10~kPa). The resulting force–displacement curves quantify the energy required for detachment (detachment work).

Although all four shapes exhibit similar maximum attachment forces, their detachment work differs markedly. Shape~1 requires substantially more energy to recover its deformation and overcome membrane–substrate adhesion, owing to its favorable flexibility and improved conformability. Consequently, Shape~1, which demonstrates the highest detachment work, is chosen for subsequent optimization.

\begin{figure*}[!ht]
    \centering
    \includegraphics[width=\linewidth]{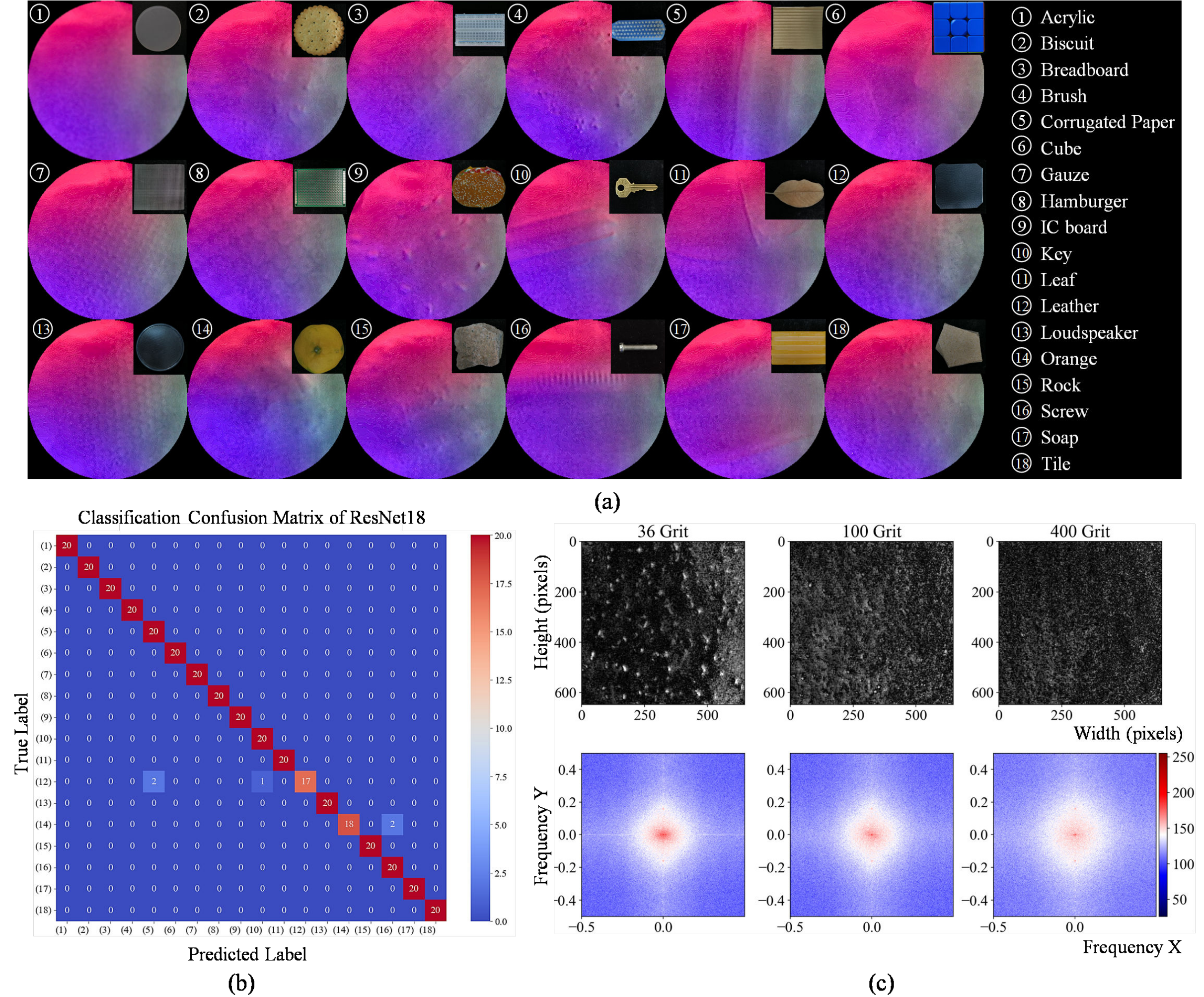}
    \caption{SuckTac perception experiments: (a) Surface texture samples of 18 daily objects acquired by the SuckTac. (b) Confusion matrix of texture classification using a ResNet18 network. (c) Differential images of sandpaper surfaces (top row) and the corresponding frequency spectra (bottom row).}
    \label{fig:camera-based tactile}
\end{figure*}

\subsection{Corrugated Lip Design}
\label{sec:lip}

Although conventional circular suction lips exhibit a certain degree of adaptability, their performance often becomes insufficient when interacting with curved or compliant surfaces. To further enhance adhesion reliability, we introduce a structural modification to the sucker lip [Fig.~\ref{fig:improvement}(c)]. The lip contour is parameterized by the function:
\begin{equation}
\begin{aligned}
    x(s) &= \big(R_0 + A \sin(Ns)\big)\cos(s), \\
    y(s) &= \big(R_0 + A \sin(Ns)\big)\sin(s),
\end{aligned}
\end{equation}
where $s$ is the angular parameter, $N$ denotes the number of sinusoidal cycles, $R_0 = 28.5~\text{mm}$ is the baseline radius, and $A = 1.5~\text{mm}$ is the amplitude. Four suckers are fabricated with $N = 0, 6, 8,$ and $10$, while keeping the inner diameter $d = 40$~mm identical.

The suction performance is evaluated on sinusoidal substrates of varying wavelength, described by 
\begin{equation}
y(x) = 0.5 \cos\left(\frac{2\pi}{T} x\right),
\end{equation}. 

To systematically evaluate adhesion performance, we test four substrate conditions with $D/T = 0, 1, 2,$ and $4$, where $D=60$~mm is the maximum outer diameter of the corrugated lip. During each test, all suckers are subjected to the same preload (5~N) and negative pressure ($-10$~kPa). On relatively flat surfaces, all lips achieve reliable adhesion, with the circular lip ($N=0$) performing slightly better due to continuous contact and uniform sealing. As surface curvature increases, however, the circular lip’s performance deteriorates rapidly. In contrast, corrugated lips with sinusoidal contours can locally deform, creating multiple independent contact regions that better adapt to surface irregularities. Among the tested variants, the lip with $N=8$ provides the most consistent adhesion across different substrates. Although higher corrugation numbers may further improve adhesion on substrates with extreme sinusoidal curvature, such cases are uncommon and beyond the scope of this study.

% \textbf{Structure Optimization}
% % 这部分主要写吸盘的基本设计和基于双曲线的优化方法,增大脱附功和吸附稳定性
% % 可以加一个小小的图示，并在其中加入参数的表示
% \lipsum[4]\lipsum[6]
\subsection{Lip Microstructures Design}
Although a wavy lip can effectively enhance the adaptability of suckers to highly curved surfaces, its adhesion performance remains limited on surfaces with microscopic roughness, such as sandpaper. To further enhance adhesion on rough surfaces, we introduce a sub-millimeter microstructure on the lip bottom. Specifically, circular holes with a diameter of 0.5~mm and a depth of 0.3~mm are uniformly distributed on three concentric rings with an interval of 0.6~mm [Fig.~\ref{fig:improvement}(d)]. This design serves two purposes: 1) the holes expel air under compression, thereby strengthening the negative pressure effect; and 2 the microhole array introduces local compliance, allowing the lip to better conform to fine surface irregularities and improving sealing performance.  

To systematically evaluate the effectiveness of this design, we compare the original lip without holes and three lips with different hole densities, corresponding to n=60, 120, and 180 holes per ring, respectively. Adhesion tests are performed under the same preload (5~N) and negative pressure ($-10$~kPa). The results show that all four suckers maintain stable adhesion on smooth surfaces with negligible differences.However, on increasingly rough surfaces, lips with microstructures show clear advantages over smooth lips, and these advantages become more pronounced as the hole density increases. Considering both manufacturability and performance, the configuration with 180 holes per ring is chosen as the standard lip design for subsequent experiments.

% \textbf{soft Lip Design}
% % 裙边的专业名词要请你查一下哦
% % 这部分主要写吸盘裙边的基本设计方法，加入了相关的波纹提高了对于粗糙表面的鲁棒性

\section{Experiments and Robot Applications}
\label{sec:Experiments}
% \subsection{Surface Perception Evaluations}
% % surface GT benchmarks
% % reuslts: error map
% % xyz analysis 
% % limitations
% \label{sec:geometry}
% To evaluate the performance of our geometry reconstruction pipeline, we conducted both qualitative and quantitative assessments on representative contact scenarios. Two categories of objects were selected: curved surface objects (sphere and ellipsoid) and flat surface objects (triangular prism and cylinder), covering a range of geometrical complexities.

% \begin{figure*}[!ht]
%     \centering
%     \includegraphics[width=\linewidth]{imgs/camera_based tactile.png}
%     \caption{The design of the proposed SuckTac. xxxxxxxx}
%     \label{fig:design and fabrication}
% \end{figure*}
\begin{figure*}[!ht]
    \centering
    \includegraphics[width=\linewidth]{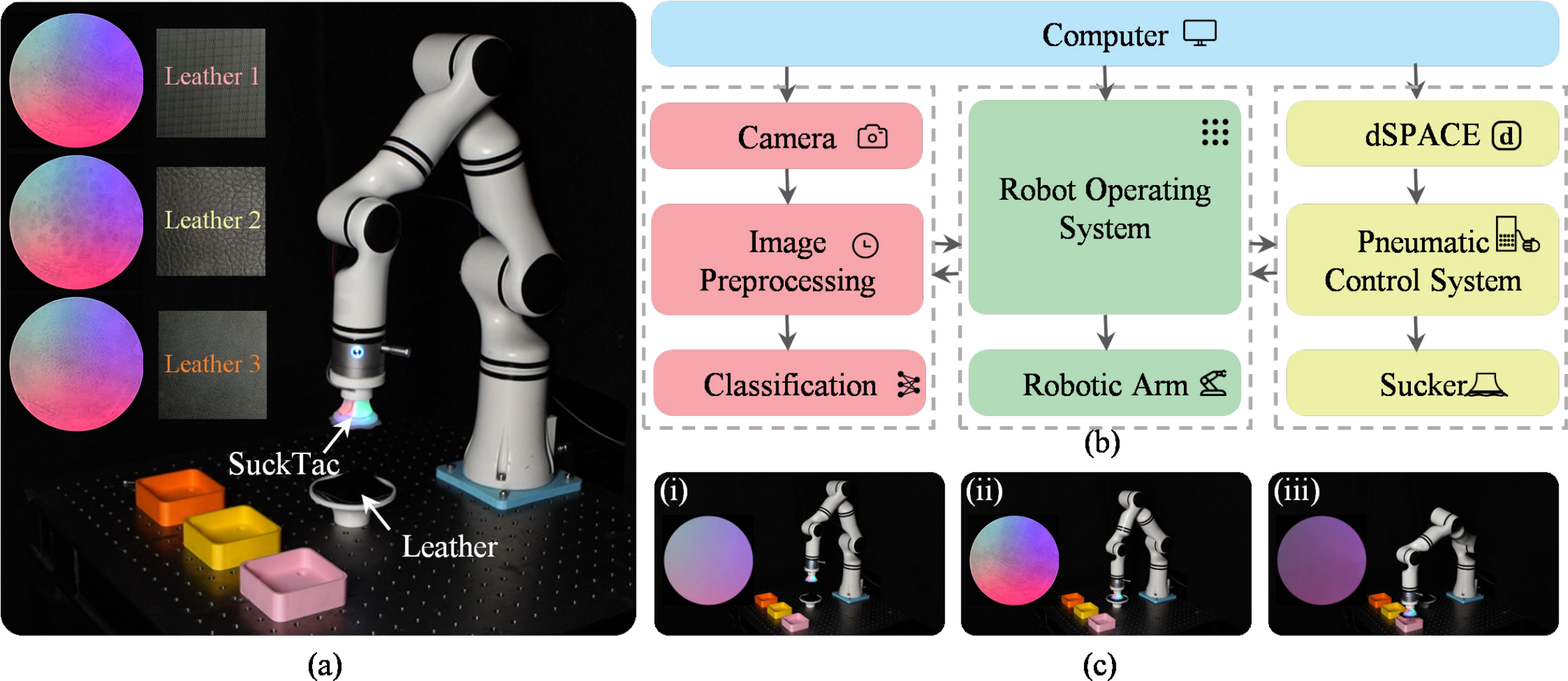}
    \caption{Grasping-classification experiments with the SuckTac: (a) Experimental setup and three leather surfaces of distinct textures. (b) System framework integrating vision, robotic arm control, and pneumatic actuation. (c) Workflow of the grasping-classification procedure.
    % \TR{ResNet18 change as classification, then take a arrow to Robot arm and sucker, for information flow! Remove this after change.}
    }
    \label{fig:grasping}
\end{figure*}

\subsection{Texture Classification Characterization}

For a tactile sensor, the ability to classify surface textures is crucial for accomplishing perception tasks. To this end, we design a camera-based tactile characterization experiment to evaluate the capability of the SuckTac in classifying surface textures, using 18 daily objects with distinct surface patterns [see Fig.~\ref{fig:camera-based tactile}(a)].In the experiment, we first collect tactile images by pressing the SuckTac on different objects under randomly sampled preload (3–5~N), pressure (5–8~kPa), and contact position, acquiring 100 images for each object. To further enhance prediction accuracy, a mask is subsequently applied to preserve only the central effective tactile region.

The tactile images are then fed into a ResNet18-based convolutional neural network for training. The network takes preprocessed $640 \times 480$ tactile images as input and outputs one-hot labels for 18 categories. Eighty percent of the images are used as the training set, while the remaining twenty percent serve as the test set to validate the network's generalization performance. Cross-entropy loss and the Adam optimizer are used for iterative training, with a learning rate of $3 \times 10^{-6}$ over 100 epochs.

The experimental results are presented as a confusion matrix [Fig.~\ref{fig:camera-based tactile}(b)], which visually illustrates the classification accuracy for each object texture and highlights frequently confused pairs. Analysis shows that the sucker achieves over 90\% tactile recognition accuracy for most objects, with slightly reduced accuracy for some pairs of visually or texturally similar objects, such as orange and brush. This observation not only reveals the inherent challenges of distinguishing between subtly varying textures, but also points to opportunities for further improvement through more advanced feature extraction or multimodal fusion. Overall, this experiment demonstrates that the SuckTac possesses strong perceptual capability in distinguishing surface textures, providing a reliable foundation for subsequent multi-object classification tasks.

% \subsection{Absorb Robustness Evaluation}
% \label{sec:force}
% 主要表征脱附功率的大小，体现稳定性
% 再表征对于不同粗糙度的适应力的大小，体现裙边的适应性；
\subsection{Surface Roughness Characterization}
%The structural design of the SuckTac passively accommodates surfaces with moderate roughness to prevent detachment, while the camera-based tactile module actively guides the SuckTac in selecting surfaces of suitable roughness, further reducing the risk of detachment. 
To quantitatively evaluate the SuckTac's perception capability across different roughness levels, we conduct a systematic characterization experiment using standard sandpapers with grit sizes ranging from 36 mesh to 400 mesh. In each trial, the SuckTac is applied under identical preload ($5~\mathrm{N}$) and positive pressure ($8~\mathrm{kPa}$), while images are simultaneously captured. To emphasize local texture variations induced by roughness, each sandpaper-contact image is processed by subtracting a reference image acquired on a smooth surface, yielding difference images that intuitively visualize microstructural distinctions [Fig.~\ref{fig:camera-based tactile}(c)]. Besides, to quantitatively assess the SuckTac's recognition capability, the difference images are subjected to a two-dimensional Fast Fourier Transform (FFT). To enhance the visibility of low-amplitude high-frequency components, logarithmic scaling is applied to the resulting spectra, providing clearer representation of texture features across different spatial scales.

%Surface roughness fundamentally corresponds to height fluctuations over spatial positions, which can be analyzed in the frequency domain. 

Experimental results show that as sandpaper grit size increases, high-frequency spectral components diminish while low-frequency components become more pronounced, highlighting the SuckTac’s ability to sense micro-scale roughness. For grit sizes below 400 mesh, frequency-domain analysis clearly distinguishes different roughness levels, whereas beyond 400 mesh the attenuated high-frequency content reduces recognition capability, largely due to the combined effects of contact mechanics and the sensor’s resolution limits. These observations define the effective sensing envelope of the SuckTac and suggest clear avenues for improvement through higher-resolution imaging or optimized contact conditions. Ultimately, the results demonstrate that the SuckTac effectively perceives surface roughness variations across a wide range, highlighting its potential to ensure reliable attachment even on complex rough surfaces through active perception.

% protential 3 apps
% grasping, shear force process
% friction estimation
% hold the cap
% claim on the coarse surface 
\begin{figure*}[!ht]
    \centering
    \includegraphics[width=\linewidth]{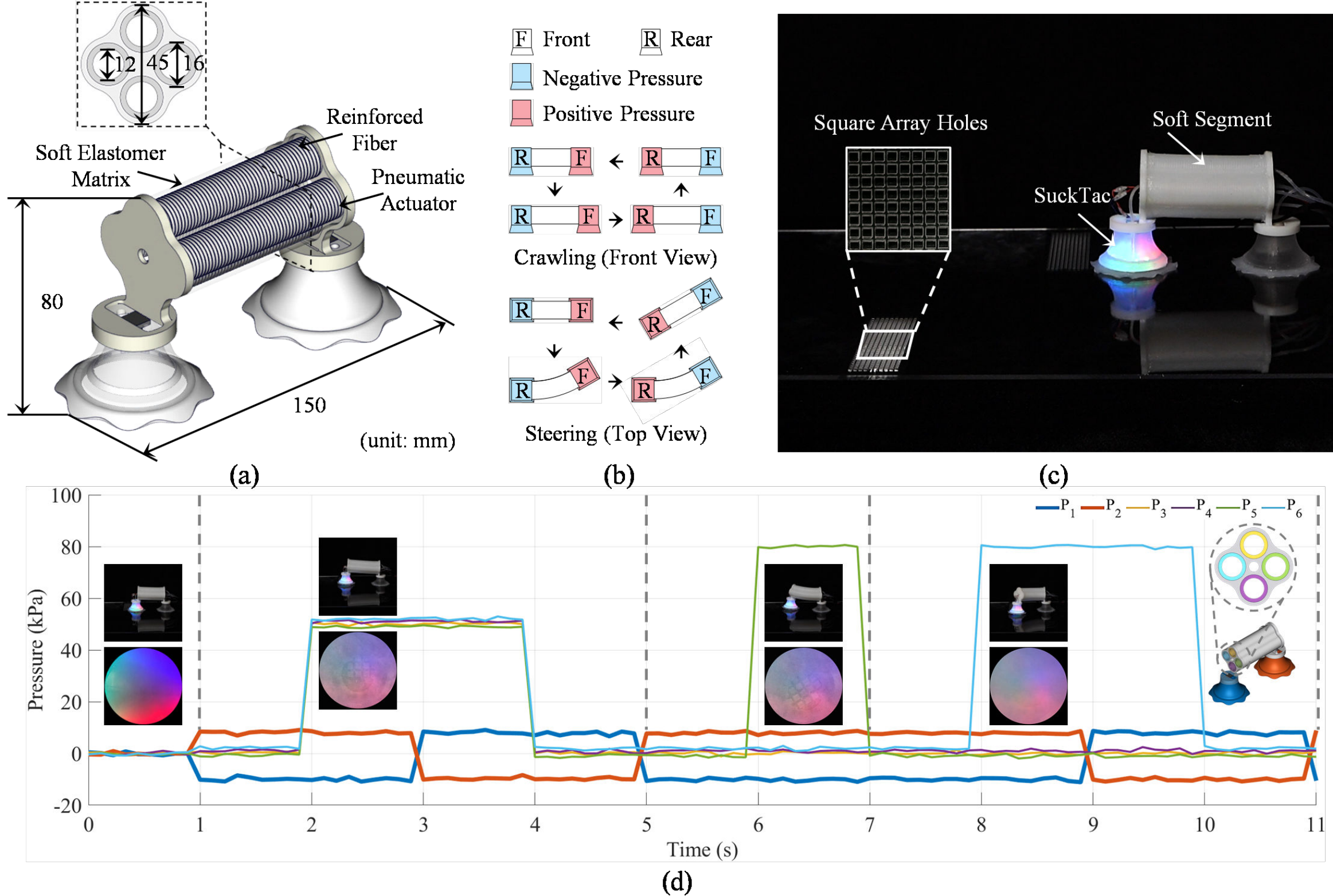}
    \caption{Locomotion perception-planning with the SuckTac-integrated soft robot. (a) Structure and dimensions of the robot. (b) Crawling and steering gaits. (c) Experimental setup with obstacle arrays. (d) Pneumatic actuation sequence and corresponding robot states.}
    \label{fig:climb_planning_equi}
\end{figure*}
\subsection{Grasping-Classification Experiment}
%Leveraging its improved structural design and integration with the camera-based tactile sensing module, the SuckTac exhibits clear advantages in grasping-classification tasks. Its compliant structure allows stable grasping of soft or rough objects that are typically challenging to handle, while the tactile module enables accurate perception and classification of the object's surface texture. 

To assess its practical performance, the SuckTac is mounted on a robotic arm end-effector and evaluated in a grasping-classification experiment involving three types of leather surfaces with distinct textures (Fig.~\ref{fig:grasping}(a)).The experimental system comprises three interconnected modules [Fig.~\ref{fig:grasping}(b)]: 1) a vision processing and classification module, where images of the sucker–leather contact are acquired by a camera, preprocessed (including cropping and normalization), and then fed into a trained ResNet18 network for real-time texture classification; 2) a robotic arm control module, in which the computer communicates with the robotic arm via ROS to execute positioning, pressing, and transport operations; and 3) a pneumatic control and actuation module, where the pneumatic system is controlled in real time through dSPACE to regulate positive and negative pressure for surface perception, reliable attachment, and release.Together, these modules form an integrated perception–action loop, where tactile sensing, precise robotic control, and dynamic pneumatic actuation work in concert to enable reliable, texture-aware grasping and classification.

The experimental procedure is illustrated in Fig.~\ref{fig:grasping}(c), with full demonstrations provided in the Supplementary Video. Under computer control, the robotic arm first moves the SuckTac close to the target leather surface, and a positive pressure of 8~kPa is applied to prepare for tactile perception. The SuckTac is then pressed onto the surface to capture contact images, which are processed by the classification network to predict the texture category. Following image classification, the pneumatic system switches to -10~kPa for firm attachment, and the robotic arm moves the object to the designated location based on the predicted category, where the SuckTac releases the leather, completing the closed-loop process of texture-aware grasping and manipulation.

This experiment demonstrate the effectiveness of the integrated tactile sensing in the sucker. Its success stems from the synergy between stable attachment provided by the suction mechanism and reliable surface classification enabled by tactile sensing, which together enable informed and adaptive grasp decisions. This capability exhibit the potential of the SuckTac for texture-aware cloth picking and manipulation in real world, where both secure attachment and accurate perception are essential.

%\subsection{Robotic tactile sensing in intravenous injection}

\subsection{Texture-aware Guided Locomotion Experiment}
%Equipped with integrated illumination, the SuckTac enables the robot to perceive surface textures even in low-light conditions, allowing it to detect and avoid obstacles during navigation. We demonstrate this capability using a soft robot with a flexible segment actuated by four fiber-reinforced pneumatic actuators, a SuckTac at the head, and a conventional suction cup at the tail. 
%怎么做的，前面方法部分已经说的非常清楚了，这里直接说为了评估什么什么的性能，直接进入实验怎么做和实验分析即可
With the integrated illumination, the SuckTac enables the robot to perceive surface textures even in low-light conditions, and to detect and avoid obstacles during navigation. We demonstrate on a soft robot with a flexible segment actuated by four fiber-reinforced pneumatic actuators, a SuckTac at the head, and a conventional suction cup at the tail. Coordinated pressure control allows the robot to perform crawling and steering gaits while actively sensing the terrain [Fig.~\ref{fig:climb_planning_equi}(a)].

As illustrated in Fig.~\ref{fig:climb_planning_equi}(b), crawling locomotion proceeds through four sequential steps. Initially, the tail sucker anchors while the head sucker is pressurized to reduce friction. Next, the body elongates forward symmetrically. This is followed by anchoring of the head sucker and release of the tail. Finally, the body contracts back to its initial configuration, completing one locomotion cycle. Steering follows the same sequence, but actuation is applied only to one lateral side to induce rotation.

During experiments [Fig.~\ref{fig:climb_planning_equi}(c)], the robot advances by default while the head sucker captures ground textures at each step. The sensor detect the surface status in movement. If no hole is present, the robot continues forward; otherwise, it checks right and left directions in sequence and turns toward the obstacle-free side before resuming straight motion. The corresponding pressure control sequence and robot states are shown in Fig.~\ref{fig:climb_planning_equi}(d), and comprehensive demonstrations are provided in the Supplementary Video.

This experiment demonstrates that the soft robot, equipped with SuckTac, can achieve autonomous obstacle avoidance on uneven surfaces using a simple perception–decision–action strategy. Such capability underscores the practical value of integrating camera-based tactile into locomotion control, enabling robots to handle environments with subtle surface variations. These findings highlight the potential of texture-aware adaptive locomotion for reliable operation in unstructured and dynamic real-world scenarios.

\section{Conclusion}
\label{sec:Conclusion}
In this paper, we introduce a smart sucker with high-density tactile perception and robust suction capabilities. By integrating a camera-based tactile sensor, we equip the sucker with high-performance tactile skills. This design not only enhances the intelligence of sucker-based manipulation and locomotion but also opens up new research directions for both tactile sensors and robotic suckers. We conduct extensive experiments to demonstrate the mechanical and perceptual performance of SuckTac, showcasing its potential in target sensing, grasping, and environmental perception for soft robots and locomotion. These experiments highlight the wide frontier for smart suckers that integrate sensing and actuation.

Although SuckTac demonstrates impressive performance, there are still some limitations. The illumination and optical path still need to be improved for more uniform illumination. Besides, the pneumatic structure and the fabrication process need to be strengthened for more stable and durable service. Furthermore, we have not fully explored the joint design between tactile perception and structure, which may yield more effective and higher-performance smart suckers.In the future, we will explore more challenging applications, integrate our system into diverse robots, and combine it with embodied intelligence for more general tasks in environmental exploration, tool operation, and household organization.
% \TR{Done.}

%\section*{Acknowledgment}
%The authors would like to thank xxx,xxx (SJTU) and xxx xxx (SJTU) for setting up the experiments;xxx (SJTU) for insightful discussions.

\ifCLASSOPTIONcaptionsoff
  \newpage
\fi

% \newpage
% \clearpage

% \end{thebibliography}
{\small
\bibliographystyle{IEEEtran}
\bibliography{refs}
}

% \end{IEEEbiography}


% % insert where needed to balance the two columns on the last page with
% %\newpage



%\vfill


\end{document}